\documentclass{article}
\usepackage{ijcai19}

\usepackage{times}
\usepackage{helvet}  
\usepackage{courier}  
\usepackage{url}  
\usepackage{booktabs}       
\usepackage{amsfonts}       
\usepackage{nicefrac}       
\usepackage{microtype}      
\usepackage{latexsym}
\usepackage{amsmath}
\usepackage{appendix}
\usepackage{amssymb}
\usepackage{authblk}
\usepackage{multirow}
\usepackage{longtable}
\usepackage{stfloats}
\usepackage{enumitem}
\usepackage{wrapfig}
\usepackage{hyperref}
\usepackage{xcolor}
\usepackage{soul}
\usepackage{graphicx, subfigure, float} 
\usepackage[ruled]{algorithm2e}
\usepackage{algorithmic}
\usepackage{color,colortbl}

\usepackage{url}
\title{GSN: A Graph-Structured Network for Multi-Party Dialogues}

\author[ ]{Wenpeng Hu$^{1,3,*}$, Zhangming Chan$^{2,3,*}$}
\author[ ]{Bing Liu$^{4,}$\thanks{Equal Contribution.}}
\author[ ]{Dongyan Zhao$^{2,3}$}
\author[1]{Jinwen Ma}
\author[ ]{Rui Yan$^{2,3,}$\thanks{Corresponding author}}
\affil[1]{Department of Information Science, School of Mathematical Sciences, Peking University}
\affil[2]{Center for Data Science, Peking University, Beijing, China}
\affil[3]{ICST, Peking University, Beijing, China}
\affil[4]{Department of Computer Science, University of Illinois at Chicago}
\affil[ ]{\textit {\{wenpeng.hu,zhangming.chan,zhaody,ruiyan\}@pku.edu.cn}}
\affil[ ]{\textit {liub@uic.edu \ \ \ \ \  jwma@math.pku.edu.cn\ \ \ \ }}

\begin{document}
\maketitle
\newcommand{\dubbelop}{$^{\blacktriangle}$}

\begin{abstract}
Existing neural models for dialogue response generation assume that utterances are sequentially organized.~However, many real-world dialogues involve multiple interlocutors (i.e., multi-party dialogues), where the assumption does not hold as utterances from different interlocutors can occur ``in parallel.''~This paper generalizes existing sequence-based models to a \textit{Graph-Structured neural Network} (GSN) for dialogue modeling.~The core of GSN is a graph-based encoder that can model the information flow along the graph-structured dialogues (two-party sequential dialogues are a special case).
Experimental results show that GSN significantly outperforms existing sequence-based models.
\end{abstract}

\section{Introduction} \label{sec:intro}

Most existing dialogue systems are sequence-to-sequence (seq2seq) models~\cite{Luan2016LSTMBC,serban2016building}.
Since a dialogue generally lasts for several turns, a dialogue session with multiple utterances can often be modeled as a sequence of ``sequences'' (i.e., utterances).
A representative framework is the hierarchical recurrent encoder-decoder framework HRED~\cite{serban2016building,Serban2017AHL}.
In HRED, a recurrent neural network (RNN) encoder encodes the tokens in each utterance, and a context RNN encodes the temporal structure of the utterances.
The entire dialogue session is then organized as a sequence.
%

\begin{figure}
  \centering
  \includegraphics[width=0.8\columnwidth]{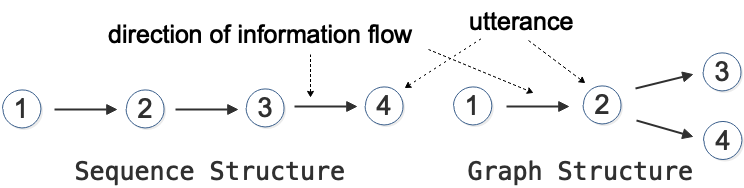}
    \caption{Sequence and graph structures.}
    \label{fig:intro_observation}
\end{figure}

Although HRED is effective in modeling sequential dialogue sessions, it falls short for dialogues involving more than two interlocutors.
Table~\ref{tab:intro} shows a real conversation of 3 people ($p_i$) in the Ubuntu forum. Utterances 3 and 4 both respond to utterance 2, represented as a graph in Figure~\ref{fig:intro_observation}.
We see that utterances can occur in parallel with each other. This is beyond the expressive power of sequence models. This paper generalizes sequence-based representation of two-party dialogues to a graph-based representation of multi-party dialogues. Two-party sequence representation is a special case. 

The proposed model, called GSN (\textit{graph-structured network}), models the information flow in a graph-structured dialogue. It is a general model and works well for both graph-structured (multi-party) and sequential (two-party) dialogues. 

%
%

\begin{table}[ht]
\renewcommand{\arraystretch}{1.3}
\scriptsize
\begin{center}
\begin{tabular}{l}
\hline
utterance 1 ($p_1$): When the screen goes blank and won't display any login page. \\
utterance 2 ($p_2$): I don't know if its a hardware problem or an os.\\
utterance 3 ($p_1$): Did you do any upgrade recently?\\
utterance 4 ($p_3$): If it works for one user it's probably not a hardware issue.\\
\hline
\end{tabular}
\end{center}
\caption{A real conversation in the Ubuntu forum.}
\label{tab:intro}
\end{table}

The core of GSN is an \textit{utterance-level graph-structured encoder} (UG-E),
which encodes utterances based on the graph topology rather than the sequence of their appearances. Encoding in UG-E is an iterative process.
In each iteration, each utterance (a node in the graph) $i$ accepts information from all its preceding utterances (nodes) $j$. 
UG-E is thus a generalization of existing sequential encoders, and can handle both sequential and graph-based dialogues.

GSN also models the speaker information as the utterances from the same speaker often have certain relationships.
Sequence-based methods in~\cite{Li2016APN,Zhang2017NeuralPR} also learn a user embedding and concatenate it to the utterances.
However, GSN builds implicit connections between utterances from the same interlocutor to model the dynamic information flow among his/her utterances with no explicit user representation, which results in performance gains. 

In summary, this paper makes the following contributions.
(1) It proposes a novel graph-structured network (GSN) to model graph-structured dialogues.
Sequence models are a special case. The core of GSN is an utterance-level graph-structured encoder (UG-E). To our knowledge, no work on graph-based representation learning has been done for dialogues.
(2) It formulates the linkage within the graph to model users across dialogue sessions. Experiments show that GSN can reach up to 13.85 BLEU points and improve the state-of-the-art baselines by 2.27 (over 16\%) BLEU points.

\section{Problem Formulation} \label{sec:ProblemFormulation}

Utterances in a structured dialogue session
can be formulated as a directed graph $\mathbf{G}(V, E)$, where $V$ is a set of $m$ vertices $\{1, ..., m\}$ and $E = \{e_{i,j}\}^m_{i,j=1}$ is a set of directed edges. Each vertex $i$ is an utterance represented as a vector $\mathbf{s}_i$ learned by an RNN. If utterance $j$ is a response to utterance $i$, then there is an edge from $i$ to $j$ with $e_{i,j} = 1$; otherwise $e_{i,j} = 0$.
Our goal is to generate the (best) response $\mathbf{\bar{r}}$ that maximizes the conditional likelihood given the graph $\mathbf{G}$:
\begin{small}
\begin{equation}\label{eq:ns1}
 \mathbf{\bar{r}} = \mathop{\arg \max}_\mathbf{r} \log P(\mathbf{r|G}) = \mathop{\arg \max}_\mathbf{r} \sum_{i=1}^{|\mathbf{r}|} \log P(r_i | \mathbf{G}, \mathbf{r}_{<i})
\end{equation}
\end{small}
where $P(\mathbf{r}|\mathbf{G})$ is modeled with the proposed GSN.

This model can be further enhanced by considering the speaker information, which introduces an adjacency matrix $U = \{u_{i,j}\}^m_{i,j=1}$, with $u_{i,j} = 1$ if utterances $i$ and $j$ are from the same speaker and $j$ is after $i$; $u_{i,j} = 0$ otherwise. 

\section{Graph-Structured Neural Network (GSN)}\label{sec:model}

\begin{figure*}[ht]
  \centering
  \includegraphics[width=0.75\textwidth]{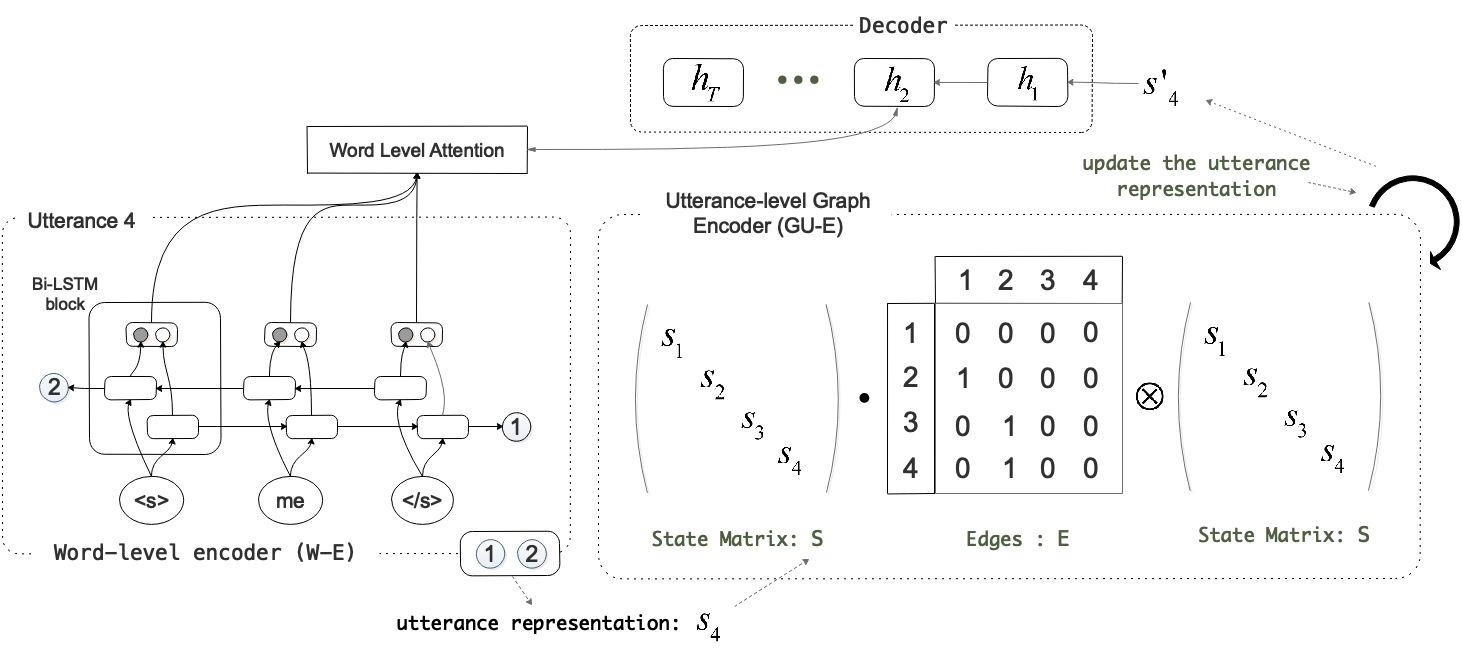}
    \caption{Architecture of GSN. }
    \label{fig:NGM_all_frame}
\end{figure*}
Figure \ref{fig:NGM_all_frame} gives the overall framework of GSN, which has three main components: a \textit{word-level encoder} (W-E), an \textit{utterance-level graph-structured encoder} (UG-E), and a decoder. UG-E is the core of GSN.
To make Figure \ref{fig:NGM_all_frame} concise, we omitted some connecting lines and attentions. 
`$\otimes$' is a special multiplication operator, called the \textit{update operator} (see below). `$\cdot$' denotes the mathematical matrix multiplication. 

\subsection{Word-level Encoder (W-E)}

Given an utterance $i  = (w_{i,1}, w_{i,2}, ..., w_{i,n})$, W-E encodes it into an internal vector representation.
We use a bidirectional recurrent neural network (RNN) with LSTM units 
to encode each word $w_{i,t}, t \in \{1, ..., n \}$ as a hidden vector $ \mathbf{s}_{i,t}$:

\begin{small}
\begin{equation} \label{eq:ns2}
\begin{aligned}
\overrightarrow{\mathbf{s}_{i,t}} =& \overrightarrow{LSTM}(\mathbf{e\_w}_{i,t}, \overrightarrow{\mathbf{s}_{i,t-1}}) \\
\overleftarrow{\mathbf{s}_{i,t}} =& \overleftarrow{LSTM}(\mathbf{e\_w}_{i,t}, \overleftarrow{\mathbf{s}_{i,t-1}})
\end{aligned}
\end{equation}
\end{small}
where $\mathbf{e\_w}_{i, t}$ is the embedding of word $w_{i,t}$ at time step $t$, $\overrightarrow{\mathbf{s}_{i,t}}$ is the hidden state for the forward pass LSTM and $\overleftarrow{\mathbf{s}_{i,t}}$ for the backward pass. We use their concatenation, i.e., $[\overrightarrow{\mathbf{s}_{i,t}}; \overleftarrow{\mathbf{s}_{i,1}}]$, as the hidden state $\mathbf{s}_{i,t}$ at time $t$. Note that each word in the utterance indicates a state and a time step. 

After encoding by W-E, a session with utterances $\{1, ..., m\}$ is represented with $\mathbf{S} = \{\mathbf{s}_i, i \in \{1, ..., m\}\}$, where $\mathbf{s}_i = \mathbf{s}_{i,n}$ is the last hidden state of W-E.

\subsection{Utterance-level Graph-Structured Encoder (UG-E)} \label{ssec:IFU}
The HRED model is a hierarchical sequence-based word and utterance-level RNN. It predicts the hidden state of each utterance at time step $t$ by encoding the sequence of all utterances appeared so far. 
Due to graph structures in real dialogues, RNN is no longer suitable for modeling the information flow of utterances.
For instance, in Figure \ref{fig:intro_observation}, HRED cannot handle utterances 3 and 4 properly because they are not logically sequential, but are ``in parallel.'' The UG-E comes to help. 


\subsubsection{UG-E \& Information Flow Over Graph} \label{ssec:IFOG}
To model a graph structure and its information flow, we propose a new RNN with dynamic iterations. 
Given a session $\mathbf{S}$, only the information in the preceding nodes/vertices $i'$ of each node $i$ will flow to $i$ in an iteration (i.e., there is a directed edge from each $i'$ to $i$). Then the state of node (utterance) $i$ is updated and the updated state is used in the next iteration. In each iteration, all updates in a session are done in parallel. In this way, the encoding information and gradients can flow fully over the graph after some iterations. For instance, in the session in Figure \ref{fig:intro_observation} (or \ref{fig:user_info_flow}), although the information flow of one iteration is from one node's preceding nodes to the node, the information in 1 can flow to 3 after two iterations.

\begin{figure}[h]
\centering
\includegraphics[width=0.4\textwidth]{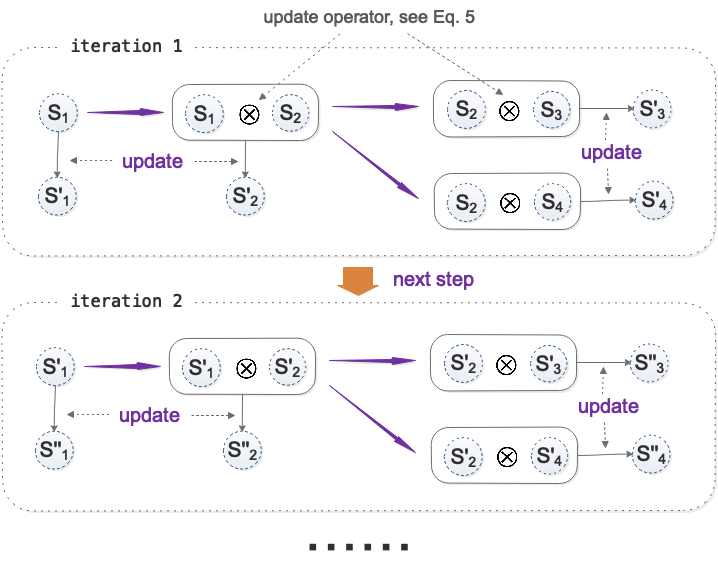}
\caption{Information flow over the graph.}
\label{fig:info_flow_step}
\end{figure}

UG-E's basic operation is illustrated in Figure \ref{fig:info_flow_step}. For example, given a session $\mathbf{S} = (\mathbf{s}_{1}, \mathbf{s}_{2}, \mathbf{s}_{3}, \mathbf{s}_{4})$, in the $l$-th iteration, the state of the $i$-th utterance can be calculated by:

\begin{small}
\begin{equation} \label{eq:ns3}
\begin{aligned}
 & \mathbf{s}^l_i = \mathbf{s}^{l - 1}_i + \eta \cdot \Delta \mathbf{s}^{l - 1}_{I|i} \\
 & \Delta \mathbf{s}^{l - 1}_{I|i} = \sum_{i' \in \varphi} \Delta \mathbf{s}^{l - 1}_{i'|i}
\end{aligned}
\end{equation}
\end{small}
where $\varphi$ is the collection of preceding nodes of the current node $i$ in the direction of the information flow; $\Delta \mathbf{s}^{l - 1}_{I|i}$ is the updating information, which is calculated by Eq. \ref{eq:ns5} below; $\eta$ is the updating coefficient indicating how much the new information (from the preceding nodes) should be added to the current state of the $i$-th utterance (node). Inspired by~\cite{sabour2017dynamic}, we design an alpha-weight as the updating coefficient. {We use a non-linear ``squashing'' function (i.e., $\text{SQH}(\cdot)$) to give vectors with a small norm a weight close to $\alpha$, but a large norm a weight close to 1:} 

\begin{small}
\begin{equation} \label{eq:ns4} 
\begin{aligned}
\eta = \text{SQH}(\Delta \mathbf{s}^{l - 1}_{I|i}) = \frac{\alpha + ||\Delta \mathbf{s}^{l - 1}_{I|i}||}{1 + ||\Delta \mathbf{s}^{l - 1}_{I|i}||}
\end{aligned}
\end{equation}
\end{small}
where $\alpha > 0$ is a hyperparameter (it should be greater than 0 to provide enough updating rate from the very beginning); $\Delta \mathbf{s}^{l - 1}_{I|i}$ is the updating information and is produced based on the state of the current utterance $\mathbf{s}^{l - 1}_i$ and the state of the preceding utterance  $\mathbf{s}^{l - 1}_{i'}$:

\begin{small}
\begin{equation} \label{eq:ns5}
\begin{aligned}
\Delta \mathbf{s}^{l - 1}_{i'|i} = \mathbf{s}^{l - 1}_{i'} \otimes \mathbf{s}^{l - 1}_{i}
\end{aligned} 
\end{equation}
\end{small}
where `$\otimes$', the \textit{update operator}, computes the updating information. Inspired by the updating operation hidden in Gated Recurrent Units (GRU)~\cite{cholearning}, $\otimes$ is defined as: 

\begin{small}
\begin{equation} \label{eq:ns6}
\begin{aligned}
\Delta \mathbf{s}^{l - 1}_{i'|i} =&\ (1 - \mathbf{x}_i) * \mathbf{s}^{l - 1}_{i'} + \mathbf{x}_i * \tilde{\mathbf{h}}_i \\\tilde{\mathbf{h}}_i =&\ \tanh(\mathbf{W} \cdot [\mathbf{r}_i * \mathbf{s}^{l - 1}_{i'}, \mathbf{s}^{l - 1}_{i}]) \\
\mathbf{x}_i =&\ \sigma(\mathbf{W}_x \cdot [\mathbf{s}^{l - 1}_{i'}, \mathbf{s}^{l - 1}_{i}] \\
 \mathbf{r}_i =&\ \sigma(\mathbf{W}_r \cdot [\mathbf{s}^{l - 1}_{i'}, \mathbf{s}^{l - 1}_{i}]
\end{aligned}
\end{equation}
\end{small}
where $\mathbf{W}$, $\mathbf{W}_x$ and $\mathbf{W}_r$ are parameters to be learned. $\sigma$ is the sigmoid function.

\subsubsection{Bi-directional Information Flow}\label{ssec:BIF}
\begin{figure}[t]
\centering 
\subfigure[\scriptsize Bi-directional information flow.]
{ 
\label{fig:info_flow_unit} 
\includegraphics[width=0.2\textwidth]{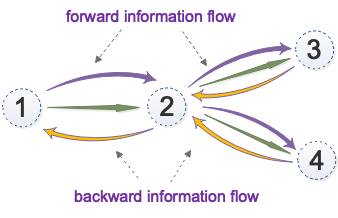}
} 
\subfigure[\scriptsize Speaker information modeling.]
{ 
\label{fig:user_info_flow}
\includegraphics[width=0.2\textwidth]{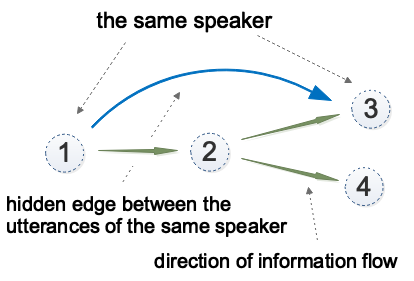}
} 
\caption{Information flow.} 
\label{fig:attn_hm} 
\end{figure}
In Figure \ref{fig:info_flow_unit}, utterances 3 and 4 are two responses to utterance 2. It is obvious that utterance 2 can help generate a better state for utterance 4 and vice versa. However, the algorithm introduced above only allows the information and gradients to flow over the forward direction of the graph (as shown by the purple arrows in Figure \ref{fig:info_flow_unit}). Hence the information in utterance 3 cannot flow to utterance 4. 

To tackle this problem, we propose a Bi-directional Information Flow (BIF) algorithm, which also uses backward information flow (as shown by the orange arrows in Figure \ref{fig:info_flow_unit}). {In order to allow information to flow thoroughly, we push the information to flow backward first and then forward to ensure that the information can flow from one node to one's sibling nodes, i.e., backward to parent and forward to siblings.} 
In our example above, the information of utterance 3 can flow to utterance 4 through utterance 2 after one backward flow and one forward flow, illustrated in Figure~\ref{fig:info_flow_unit}. 

\subsubsection{Speaker Information Flow} \label{ssec:userinfoflow}

Representing speaker information in the latent embedding space is a popular method to enhance dialogue generation. However, this method lacks the ability to model the speaker and the dynamic changes of the speaker's ideas in a given session, especially when the speaker only speaks a few times because there may not be enough data to train those embeddings to represent the speaker and the changes since this method usually requires large data to train~\cite{Li2016APN,Qian2017AssigningPT,Zhang2017NeuralPR}. 


Since the changes in speaker utterances reflect the changes in his/her mind, 
we propose to create an edge for every pair of utterances from the same speaker following the chronological order of the utterances. Thus there should be hidden edges among all utterances of the same user (e.g., the edge from utterances 1 to 3 in Figure \ref{fig:user_info_flow}). We employ the same $\otimes$ operation to process the hidden edges, but due to different parameters, we use $\circledast$ to denote it:

\begin{small}
\begin{equation} \label{eq:ns7}
\begin{aligned}
\Delta \mathbf{s'}^{l - 1}_{i'|i} = \mathbf{s}^{l - 1}_{i'} \circledast \mathbf{s}^{l - 1}_{i}
\end{aligned}
\end{equation}
\end{small}
We add the speaker information to Eq. \ref{eq:ns3}:

\begin{small}
\begin{equation} \label{eq:ns8}
\begin{aligned}
\mathbf{s}^l_i = &\mathbf{s}^{l - 1}_i + \eta \cdot \Delta \mathbf{s}^{l - 1}_{I|i} + \lambda \cdot \Delta \mathbf{s'}^{l - 1}_{I|i} \\ 
& \Delta \mathbf{s'}^{l - 1}_{I|i} = \sum_{i' \in \varphi} \Delta \mathbf{s'}^{l - 1}_{i'|i}
\end{aligned}
\end{equation}
\end{small}
where $\eta$ and $\Delta \mathbf{s}^{l - 1}_{I|i}$ are the same as those in Eq. \ref{eq:ns3}; $\lambda$ is also calculated with Eq. \ref{eq:ns4} with input $\Delta \mathbf{s'}^{l - 1}_{I|i}$ instead of $\Delta \mathbf{s}^{l - 1}_{I|i}$.
%
\subsection{Reformulation as Matrix Operations} 
\label{ssec:CIFU}
So far, we have presented the proposed model. For computation, we reformulate it as matrix operations (also see UG-E in Figure~\ref{fig:NGM_all_frame}) and give the pseudo-code in Algorithm 1.
Recall the session $\mathbf{S} = (\mathbf{s}_{1}, \mathbf{s}_{2}, \mathbf{s}_{3}, \mathbf{s}_{4})$, which is used to build the graph $\mathbf{G}(V,E)$ in Figure \ref{fig:info_flow_unit}. We build a state matrix $\mathbb{S}$ with the vertices of the graph $\mathbf{G}$ (also the session $\mathbf{S}$) as the diagonal elements, and all the other elements are set to 0 (we name this process the \emph{Building State Matrix function}, denoted by $BSM(\mathbf{S})$). We then use the ``@'' relation (a speaker responding to another speaker) as the connection between two vertices to build the edge matrix $\mathbf{E}$ (shown in Figure \ref{fig:NGM_all_frame}). Recall the  speaker information modeling in Section \ref{ssec:userinfoflow} and the utterance speaker adjacency matrix $U$. The main operation of Eq. \ref{eq:ns8} can be formalized by:

\begin{small}
\begin{equation} \label{eq:ns9}
\begin{aligned}
\Delta \mathbf{E} = \mathbb{S}^{l - 1} \cdot \mathbf{E} \otimes \mathbb{S}^{l - 1}; \Delta \mathbf{U} = \mathbb{S}^{l - 1} \cdot \mathbf{U} \circledast \mathbb{S}^{l - 1}
\end{aligned}
\end{equation}
\begin{equation} \label{eq:ns10}
\begin{aligned}
\mathbb{S}^l = \mathbb{S}^{l - 1} + BSM(\boldsymbol{\eta} \odot \Delta \mathbb{E} + \boldsymbol{\lambda} \odot \Delta \mathbb{U})
\end{aligned}
\end{equation}
\end{small}
where $\Delta \mathbb{E} = \{\sum^m_{j=1} \Delta \mathbf{E}_{i,j}\}^m_{i=1}$ and $\Delta \mathbb{U} = \{\sum^m_{j=1} \Delta \mathbf{U}_{i,j}\}^m_{i=1}$ are two vectors, $m$ is the length of the given session; $\odot$ denotes the Hadamard product; $\boldsymbol{\eta}$ and $\boldsymbol{\lambda}$ can be calculated by:

\begin{small}
\begin{equation} \label{eq:ns11}
\begin{aligned}
\boldsymbol{\eta} = \{\text{SQH}(\Delta \mathbb{E}_{i})\}^m_{i=1}; \boldsymbol{\lambda} = \{\text{SQH}(\Delta \mathbb{U}_{i})\}^m_{i=1}
\end{aligned}
\end{equation} 
\end{small}
This is just the forward information flow. We can obtain the backward information flow operation by changing Eq. \ref{eq:ns9}:

\begin{small}
\begin{equation} \label{eq:ns12}
\begin{aligned}
\Delta \mathbf{E} = \mathbb{S}^{l - 1} \cdot \mathbf{E}^T \otimes \mathbb{S}^{l - 1}; \Delta \mathbf{U} = \mathbb{S}^{l - 1} \cdot \mathbf{U}^T \circledast \mathbb{S}^{l - 1}
\end{aligned}
\end{equation}
\end{small}
To obtain $\Delta \mathbb{E}$ and $\Delta \mathbb{U}$ in Eq.~\ref{eq:ns10}, we need to change the direction of the sum, i.e., $\Delta \mathbb{E} = \{\sum^m_{j=1} \Delta \mathbf{E}_{j,i}\}^m_{i=1}$ and $\Delta \mathbb{U} = \{\sum^m_{j=1} \Delta \mathbf{U}_{j,i}\}^m_{i=1}$.
$\mathbb{S}, \mathbb{E}, \text{and}\ \mathbb{U}$ can be very sparse. But the proposed method can be well organized and the sparse matrices can be addressed by sparse matrix operations. The pseudo-code is given in Algorithm 1 in \textit{Appendix}~\footnote{\href{https://morning-dews.github.io/Appendix/IJCAI2019_GSN.pdf}{https://morning-dews.github.io/Appendix/IJCAI2019\_GSN.pdf}}.

\subsection{Decoder}\label{ssec:decoder}
As shown in Figure \ref{fig:NGM_all_frame}, we illustrate a session $\{i\}^m_{i=1}$ with the corresponding encoding state denoted by $\mathbf{S}$. To generate a response to an utterance $i$, the decoder calculates a distribution over the vocabulary and sequentially predicts word $r_k$ using a softmax function:

\begin{scriptsize}
\begin{equation} \label{eq:ns13}
p(\mathbf{r}|\mathbf{S}; \theta) = \prod^{|\mathbf{r}|}_{k=1} P(r_k|\mathbb{S}_{i,i}, \mathbf{r}_{<k}; \theta) = \prod^{|\mathbf{r}|}_{k=1} \text{softmax}(f(\mathbf{h}_k, \mathbf{c}_k, r_{k-1}))
\end{equation}
\end{scriptsize}
where $f(\cdot)$ is the tanh function.~$r_{k-1}$ is the word generated at the (\textit{k}-1)-th time step, obtained from a word look-up table. 
$\mathbf{h}_k = \text{GRU}(\mathbf{h}_{k-1}, r_{k-1})$ is the hidden state variable of a GRU at time step $k$. $\mathbf{h}_0 = \mathbb{S}_{i,i}$, and $\mathbf{c}_k$ is the attention-based encoding of utterance $i$ at decoding time step $k$ and it is calculated by 
$\mathbf{c}_k = \sum_{j=1}^n \frac{\text{exp}(e_{j,k}) \mathbf{s}_{i,j}}{\sum_{j = 1}^{m}\text{exp}(e_{j,k})}$,
where $\mathbf{s}_{i,j}$ is the encoder hidden state at time step $j$ for utterance $i$, and $e_{j,k} = \mathbf{h}_k \mathbf{W}_a \mathbf{s}_{i,j}$ scores the match degree of $\mathbf{h}_k$ and $\mathbf{s}_{i,j}$. 

\section{Experiments}

\begin{table*}[t!]
\renewcommand{\arraystretch}{1.2}
\scriptsize
\begin{center}
\begin{tabular}{|l|cccc|c|c|ccc|} 
\hline
Model & BLEU 1 & BLEU 2 & BLEU 3 & BLEU 4 & METEOR & $\text{ROUGE}_\text{L}$ \\
\hline
seq2seq & 10.45 & 4.13 & 2.08 & 1.02 & 3.43 & 9.67  \\
seq2seq W-speaker & 10.70 & 4.98 & 2.20 & 1.55 & 3.92 & 9.42  \\
Seq2seq (last utte) & 9.85 & 3.04 & 1.38 & 0.67 & 3.98 & 8.34 \\
HRED~\cite{serban2016building} & 10.80 & 4.60 & 2.54 & 1.42 & 4.38 & 10.23 \\
\rowcolor{lightgray}
HRED W-speaker & 11.23 & 4.82 & 3.06 & 1.64 & 4.36 & 10.98 \\
\hline
GSN No-speaker (1-iter) & 9.42 & 3.05 & 1.61 & 0.95 & 3.74 &  7.63 \\
GSN No-speaker (2-iter) & 12.06 & 4.87 & 2.80 & 1.70 & 4.32 & 10.09  \\
GSN No-speaker (3-iter) & 12.77\dubbelop & 5.37\dubbelop & 3.17 & 1.99\dubbelop & 4.53 & 10.80 \\
\hline
GSN W-speaker (1-iter) & 10.31 & 4.06 & 2.34 & 1.45 & 3.88 & 9.96 \\
GSN W-speaker (2-iter) & 12.77 & 4.93 & 2.61 & 1.46 & 4.79 & 11.34 \\
GSN W-speaker (3-iter) & \textbf{\underline{13.50}}\dubbelop & \textbf{\underline{5.63}}\dubbelop & \textbf{\underline{3.24}}\dubbelop & \textbf{\underline{1.99}}\dubbelop & \textbf{\underline{4.85}}\dubbelop & \textbf{\underline{11.36}}\dubbelop \\
\hline
\end{tabular}
\end{center}
\caption{Experimental results, conducted in different settings, including sequential data and graph data using different models based on automated evaluation. 'Seq2seq (last utte)' is trained by using only the last utterance before the final response of the session as the input (all utterances before are ignored). `$n$-iter' means that the results are obtained after $n$ iterations. `No-speaker' is our proposed GSN model without speaker information flow while `W-speaker' has it. \dubbelop denotes the $p$-value $< 0.01$ in paired $t$-test against the best baseline (shaded row). 
}
\label{tab:auto_eval}
\end{table*}

\subsection{Experimental Setups} \label{ssec:dataprepara}

\textbf{Data Preparation}: Our experiment uses the Ubuntu Dialogue Corpus\footnote{http://dataset.cs.mcgill.ca/ubuntu-corpus-1.0/}~\cite{Lowe2015TheUD} as it is the only benchmark corpus with annotated multiple interlocutors.~It is also large with almost one million multi-turn dialogues, over seven million utterances and 100 million words. Each record contains a response utterance with its speaker ID and posting time.

To build the training and testing datasets, we extract all utterances with response relations indicated by the ``@'' symbol in the corpus. For example, ``A @ B'' means that the utterance is addressed to Speaker B by Speaker A. Utterances from Speaker A and Speaker B are encoded into vector representations and used to construct the state matrix introduced in Section \ref{ssec:CIFU} as vertices. A directed edge is installed from A to B and used to build the edge matrix described in Sec. \ref{ssec:CIFU}. 

Following the baselines, we take the last utterance in each given session as the utterance to be generated\footnote{GSN can generate responses all the way through a dialogue as GSN can gather the dialogue information to any graph node with the help of the dynamic iteration and information flow mechanisms.} (i.e., the output target) and the other utterances in the session as the input. Finally, we extracted 380k sessions (about 1.75M utterances) as the experiment corpus and each session has 3 to 10 utterances and 2 to 7 interlocutors. We randomly divide the corpus into the training, development (with 5k q/a pairs), and test (with 5k q/a pairs) sets. We report the results on the test set. In testing, following the graph structure, the system knows which utterances to respond to. This is reasonable as this is also the case in a human dialogue, i.e., before responding, we know which preceding utterances to rely to. 

It is important to note here that GSN is a general model that works well for both graph-structured (multi-party) and sequential (two party) dialogues as we will see shortly.

\noindent \textbf{Baselines}:
We use a seq2Seq model and a hierarchical or HRED model as the baselines. HRED has been shown superior to other state-of-the-art models~\cite{serban2016building,Sutskever2014SequenceTS}. 

\textbf{HRED model:} We use HRED~\cite{serban2016building}, the latest HRED model. Since HRED cannot deal with structured sessions, we convert each graph-structured dialogue session to a set of dialogue sequences by extracting every possible forward path from the first utterance to the last utterance of the session by following the ``@'' relationship and regarding each path (a sequence) as a dialogue sequence/session. 

\textbf{Seq2Seq model:} We use the same sequential sessions extracted for HRED. Since the last utterance in each new session is the target utterance to be generated and the others are the input utterances, all the input utterances are concatenated into a \textit{long} utterance as the input. Seq2seq modeling with attention is performed as in~\cite{Bahdanau2014NeuralMT,Sutskever2014SequenceTS} on the concatenated utterance.

\textbf{Baselines with speaker information:} We tried 3 methods of adding speaker information to baselines, using trainable speaker embedding (SE)~\cite{Li2016APN,Qian2017AssigningPT} as speaker information: a) concatenating SE with each word in the utterance; b) using SE as the prefix of the utterance; c) using SE as the prefix of the utterance and addressee’s SE as the suffix. For seq2seq (respectively, HRED), c) (b)) is the best, we report the corresponding results in Table \ref{tab:auto_eval}.

\textbf{Our GSN Model}: There are multiple ways to implement the proposed GSN model. Since the objective of this paper is not to explore all possible ways,  
we use GRU units for all recurrent neural networks. For fairness, we adopt this setting for all baselines as well. We set $\alpha$ in Eq. \ref{eq:ns4} to 0.25. 


Benefited from dynamic iterations, GSN is very flexible in generating responses for any utterance in a given session. However, to be consistent with the baselines, only the last utterance in each session is used as the target in training and testing. The code of our model can be found here \footnote{\href{https://github.com/morning-dews/GSN-Dialogues}{https://github.com/morning-dews/GSN-Dialogues}}.

\noindent \textbf{Training Details of Our GSN Model}: 
We share the word embedding between the word-level encoder and the decoder and limit the shared vocabulary to 30k. The number of hidden units is set as 300 and the word embedding dimension is set as 300. We have 2 layers for both word-level encoder and decoder.
The network parameters are updated using the Adam algorithm~\cite{Kingma2014AdamAM} with the learning rate of 0.0001. All utterances are clipped to 30 words. We run all experiments on a single GTX Titan X GPU, and training takes 25 epochs. Each experiment takes about 48 hours.

\subsection{Results and Analysis} \label{ssec:resultanalysis}
\textbf{Automated Evaluation}: We use two kinds of metrics in automated evaluation: 1) Following \cite{Fu2017AligningWT,Havrylov2017EmergenceOL}, 
we use the evaluation package of~\cite{Chen2015MicrosoftCC}, which includes BLEU 1 to 4, 
METEOR
and $\text{ROUGE}_\text{L}$.
2) We also use embedding-based metrics~\cite{forgues2014bootstrapping} which can cover the weaknesses of the BLEU's.

Table \ref{tab:auto_eval} shows the evaluation results. The first three rows are for the baselines. The three rows in the middle are for our GSN model using only the information flow over the graph structure, and the last three rows are also for our GSN model but with the addition of the speaker information flow. From Table~\ref{tab:auto_eval}, we can make the following observations:

(1).~GSN (row 11, with the speaker information flow after 3 iterations) markedly outperforms the baselines (rows 2 and 5) by up to 2.27 BLEU points (BLEU 1). 

(2).~With-speaker (W-speaker) versions of GSN also clearly outperform the no-speaker versions, indicating the importance of the speaker information flow. To further verify whether the improvement is due to adding more connections or adding the speaker edges, we conducted experiments by adding some random edges with different percentages until full connections among nodes (using No-speaker setting with 3 iterations). The results showed a clear drop with the increase of randomly added edges and received very poor result for full connection, which further shows the usefulness of the proposed speaker information modeling method. As the results are very poor, they are not shown here. 

{(3).~We also use the exist embedding-based persona method to arm baselines (W-speaker) for a fair comparison with GSN W-speaker version. We can see from Table \ref{tab:auto_eval} the gain is limited, and our method still outperforms the baselines.}

(4).~The results of GSN improve as the number of iterations increases, which indicates the importance of the dynamic iterations. With more iterations, more utterances will be modeled by GSN. Only after two iterations, our models with the speaker information flow (row 10) already outperforms both two baselines in 4 out of 6 evaluation metrics. Even for the no-speaker versions (row 7), our model beats the best baseline in 4 out of 6 evaluation metrics. 

{The BLEU scores had a tiny increase in the 4th iteration (around 0.1\% for GSN W-speaker, and 0.3
\% for GSN No-speaker). For other metrics, e.g., METEOR and ROUGEL, there is little change. The 5th iteration is similar, but the scores decrease from the 6th iteration. We thus choose 3 iterations.}
\begin{table*}[t!]
\renewcommand{\arraystretch}{1.2}
\scriptsize
\begin{center}
\begin{tabular}{|l|cccc|c|c|} 
\hline
Model & BLEU 1 & BLEU 2 & BLEU 3 & BLEU 4 & METEOR & $\text{ROUGE}_\text{L}$ \\
\hline
HRED~\cite{serban2016building} (sequential) & 9.61 & 3.48 & 1.86 & 1.01 & 4.08 & 8.22 \\
GSN No-speaker (2-iter sequential) & 11.39 & 4.55 & 2.68 & 1.71 & 4.40 & 9.74 \\
GSN W-speaker (1-iter sequential) & 8.69 & 3.1 & 1.78 & 1.19 & 3.67 & 9.19 \\
GSN W-speaker (2-iter sequential) & \underline{12.72} & 4.84 & 2.59 & 1.59 & \underline{4.70} & \underline{11.41} \\
GSN W-speaker (3-iter sequential) & 12.03 & \underline{4.92} & \underline{2.94} & \underline{1.97} & 4.31 & 10.1 \\
\hline
HRED~\cite{serban2016building} (graph) & 12.16 & 4.90 & 2.68 & 1.49 & 4.42 & 10.90 \\
GSN No-speaker (2-iter graph) & 12.35 & 5.17 & 3.08 & 1.81 & 4.43 & 10.42 \\
GSN W-speaker (1-iter graph) & 10.66 & 4.36 & 2.52 & 1.50 & 3.97 & 10.10 \\
GSN W-speaker (2-iter graph) & 12.76 & 5.23 & 2.94 & 1.75 & 4.80 & 11.33 \\
GSN W-speaker (3-iter graph) & \textbf{\underline{13.85}} & \textbf{\underline{5.83}} & \textbf{\underline{3.33}} & \textbf{\underline{1.98}} & \textbf{\underline{5.10}} & \textbf{\underline{11.66}} \\
\hline
\end{tabular}
\end{center}
\caption{Experimental results of {using \emph{sequential data} (with only 2 interlocutors, the first five rows of the results) or \emph{graph data} only (with more than 2 interlocutors, the last five rows of the results). The symbol string `$n$-iter', `No-speaker' and `W-speaker' in the table have the same meaning as those in Table \ref{tab:auto_eval}. The result of GSN No-speaker 3-iter isn't given as it performs worse.}}
\label{tab:mul-seq}
\end{table*}
\noindent \textbf{Embedding-based metrics}. Based on the embedding-based metrics, our model also outperforms the baselines. Our model gets 0.770 / 1.040 / 0.651 for the three embedding-based metrics (Embedding Average Score / Embedding Greedy Score / Embedding Extrema Score), which are all better than the scores of the best baseline model (HRED), 0.515 / 0.905 / 0.325. 
{See the details in \textit{Appendix}$^1$, 
which also includes a \textit{case study} with examples}.

\noindent \textbf{Sequential data and graph data}. To verify the generic nature of the GSN model, we conduct an ablation experiment with only sequential data (sessions with only two interlocutors) or graph data (remaining sessions with more than two interlocutors). The results are shown in Table \ref{tab:mul-seq}. We see that GSN significantly outperforms the strong HRED baseline in both sequential and graph settings. {From both Tables \ref{tab:mul-seq} and \ref{tab:auto_eval}, we can see that GSN improves in the sequential case mainly because of the additional encoding iterations.} 

{Tables \ref{tab:mul-seq} and \ref{tab:auto_eval} also show that the proposed iterative graph-structured encoder UG-E and the graph-based speaker information flow are effective. GSN is thus a good generalization of the sequence-based models, and a desirable system for both graph-structured (multi-party) and sequential (two-party) dialogue response generation.} 
\begin{table}[h]
\renewcommand{\arraystretch}{1.2}
\footnotesize
\begin{center}
\begin{tabular}{|c|c|cc|cc|}
\hline
\multirow{2}{*}{\bf Human }  & \multirow{2}{*}{{\bf HRED} } & \multicolumn{2}{c|}{\bf No-speaker } & \multicolumn{2}{c|}{\bf W-speaker } \\
\cline{3-6}
 & & 1-iter & 3-iter & 1-iter & 3-iter \\
\hline
3.01 & 1.91 & 1.89 & 1.98 & 2.23\dubbelop & \textbf{\underline{2.37}}\dubbelop\\
\hline
\end{tabular}
\end{center}
\vspace{-2mm}
\caption{Human evaluation results. 
\dubbelop denotes $p$-value $< 0.01$ in paired $t$-test against HRED. 
The perfect score is 4.}
\label{tab:human_eval}
\end{table}

\noindent \textbf{Human Evaluation}: 
We also conducted human evaluation to measure the quality of responses generated by all methods. We evaluate based on ``\textit{naturalness}'', which includes 1) grammaticality, 2) fluency and 3) rationality. We randomly sampled 100 utterance-response pairs, shuffled the order of systems, and asked three Ph.D. students to rate the pairs in terms of model quality on 0 to 4 scales (4 for the best) and we report their average scores. More details can be found in \textit{Appendix}$^1$.


Table \ref{tab:human_eval} shows that GSN (with the speaker information flow after only 1 iteration or the no-speaker version after 3 iterations) outperforms the best baseline (HRED), indicating that GSN generates more natural responses. 
The reason that our model (with the speaker information flow and just one iteration, column 5 in Table~\ref{tab:human_eval}) outperforms three iterations of our model's no-speaker version (column 4 in Table~\ref{tab:human_eval}) is because the model (with the speaker information flow) can generate a more consistent response for the speaker. The generated utterances are more preferred by humans.


\section{Related Work}
Existing dialogue models follow the sequential information flow
~\cite{Shang2015NeuralRM,Wen2017ANE,tao2019multi}.
Recent progresses in seq2seq models
~\cite{Sutskever2014SequenceTS,Luong2015EffectiveAT}
have inspired several efforts
~\cite{Li2019Insufficient,Young2017AugmentingED}
to build dialogue systems. 

Although seq2seq models have achieved good results for dialogue generation, they regard all input utterances as a long sequence, which greatly increases the complexity of the model in passing information and computing gradients.
As an improved solution, the HRED models
~\cite{serban2016building}
tackle this problem by constructing the sequential flow at the utterance level.
However, this setting is insufficient for modeling dialogues that have more than 2 interlocutors, which need a graph-based model. 

For multi-party dialogues, prior work have employed retrieval-based approaches
\cite{Zhang2017AddresseeAR,Meng2017TowardsNS}.
No graph modeling method has been proposed, although graph-based methods have been used for other NLP tasks, e.g., Graph Convolutional Networks for classification \cite{kipf2016semi} and semantic role labeling \cite{marcheggiani2017encoding}, Gated Graph Neural Networks for generation from AMR graphs and syntax-based neural machine translation \cite{beck2018graph}. Different from these works, we propose a generation model by formulating the complex dialogue problem using a graph-based solution. 
%
Also importantly, compared to seq2seq and HRED,  GSN not only can encode graph structured information flows, but also sequential ones.
%

\section{Conclusion} 
{In this paper, we proposed a \textit{general graph-structured neural network} GSN to model both graph-structured (multi-party) and sequential (two-party) dialogues. The core of the model is an utterance-level graph-based encoder (UG-E), which is a generalization of the conventional sequence-based encoder. For the response generation in multi-party conversations, the speaker information is also modeled in the graph. As our results showed, GSN is general and is suitable for both multi-party and two-party dialogues. 

The current GSN relies on clear addressee information. Our future work will try to automatically identify the conversation structure and decide who to respond to. Dynamic routine and attention can be leveraged to achieve this goal.} 

\section{Acknowledgements} 
This work was partially supported by National Key Research and Development Program of China (No.2017YFC0804001), National Science Foundation of China (No. U1604153, No. 61876196, No. 61672058), Alibaba Innovative Research 
Fund. Rui Yan was supported by CCF-Tencent Open Research Fund and Microsoft Research Asia Collaborative Research Program.

\bibliographystyle{named}
\bibliography{ijcai19}

\begin{thebibliography}{}

\bibitem[\protect\citeauthoryear{Bahdanau \bgroup \em et al.\egroup
  }{2014}]{Bahdanau2014NeuralMT}
Dzmitry Bahdanau, Kyunghyun Cho, and Yoshua Bengio.
\newblock Neural machine translation by jointly learning to align and
  translate.
\newblock {\em Computer Science}, 2014.

\bibitem[\protect\citeauthoryear{Beck \bgroup \em et al.\egroup
  }{2018}]{beck2018graph}
Daniel Beck, Gholamreza Haffari, and Trevor Cohn.
\newblock Graph-to-sequence learning using gated graph neural networks.
\newblock {\em ACL}, 2018.

\bibitem[\protect\citeauthoryear{Chen \bgroup \em et al.\egroup
  }{2015}]{Chen2015MicrosoftCC}
Xinlei Chen, Hao Fang, Tsung-Yi Lin, Ramakrishna Vedantam, Saurabh Gupta, Piotr
  Doll{\'a}r, and C~Lawrence Zitnick.
\newblock Microsoft coco captions: Data collection and evaluation server.
\newblock {\em arXiv preprint arXiv:1504.00325}, 2015.

\bibitem[\protect\citeauthoryear{Cho \bgroup \em et al.\egroup
  }{2014}]{cholearning}
Kyunghyun Cho, Bart van Merri{\"e}nboer, Caglar Gulcehre, Dzmitry Bahdanau,
  Fethi Bougares~Holger Schwenk, and Yoshua Bengio.
\newblock Learning phrase representations using rnn encoder--decoder for
  statistical machine translation.
\newblock In {\em EMNLP}, 2014.

\bibitem[\protect\citeauthoryear{Forgues \bgroup \em et al.\egroup
  }{2014}]{forgues2014bootstrapping}
Gabriel Forgues, Joelle Pineau, Jean-Marie Larchev{\^e}que, and R{\'e}al
  Tremblay.
\newblock Bootstrapping dialog systems with word embeddings.
\newblock In {\em Nips, modern machine learning and natural language processing
  workshop}, volume~2, 2014.

\bibitem[\protect\citeauthoryear{Fu \bgroup \em et al.\egroup
  }{2017}]{Fu2017AligningWT}
Kun Fu, Junqi Jin, Runpeng Cui, Fei Sha, and Changshui Zhang.
\newblock Aligning where to see and what to tell: Image captioning with
  region-based attention and scene-specific contexts.
\newblock {\em PAMI}, 39:2321--2334, 2017.

\bibitem[\protect\citeauthoryear{Havrylov and
  Titov}{2017}]{Havrylov2017EmergenceOL}
Serhii Havrylov and Ivan Titov.
\newblock Emergence of language with multi-agent games: Learning to communicate
  with sequences of symbols.
\newblock In {\em NIPS}, 2017.

\bibitem[\protect\citeauthoryear{Kingma and Ba}{2014}]{Kingma2014AdamAM}
Diederik~P Kingma and Jimmy Ba.
\newblock Adam: A method for stochastic optimization.
\newblock {\em arXiv preprint arXiv:1412.6980}, 2014.

\bibitem[\protect\citeauthoryear{Kipf and Welling}{2016}]{kipf2016semi}
Thomas~N Kipf and Max Welling.
\newblock Semi-supervised classification with graph convolutional networks.
\newblock {\em arXiv preprint arXiv:1609.02907}, 2016.

\bibitem[\protect\citeauthoryear{Kipf and Welling}{2017}]{kipf2017semi}
Thomas~N. Kipf and Max Welling.
\newblock Semi-supervised classification with graph convolutional networks.
\newblock In {\em International Conference on Learning Representations (ICLR)},
  2017.

\bibitem[\protect\citeauthoryear{Li \bgroup \em et al.\egroup
  }{2016}]{Li2016APN}
Jiwei Li, Michel Galley, Chris Brockett, Georgios Spithourakis, Jianfeng Gao,
  and Bill Dolan.
\newblock A persona-based neural conversation model.
\newblock In {\em ACL}, volume~1, pages 994--1003, 2016.

\bibitem[\protect\citeauthoryear{Li \bgroup \em et al.\egroup
  }{2019}]{Li2019Insufficient}
Juntao Li, Lisong Qiu, Bo~Tang, Dongmin Chen, Dongyan Zhao, and Rui Yan.
\newblock Insufficient data can also rock！learning to converse using smaller
  data with augmentation.
\newblock In {\em AAAI}, 2019.

\bibitem[\protect\citeauthoryear{Lowe \bgroup \em et al.\egroup
  }{2015}]{Lowe2015TheUD}
Ryan~Joseph Lowe, Nissan Pow, Iulian Serban, and Joelle Pineau.
\newblock The ubuntu dialogue corpus: A large dataset for research in
  unstructured multi-turn dialogue systems.
\newblock In {\em SIGDIAL}, 2015.

\bibitem[\protect\citeauthoryear{Luan \bgroup \em et al.\egroup
  }{2016}]{Luan2016LSTMBC}
Yi~Luan, Yangfeng Ji, and Mari Ostendorf.
\newblock Lstm based conversation models.
\newblock {\em arXiv preprint arXiv:1603.09457}, 2016.

\bibitem[\protect\citeauthoryear{Luong \bgroup \em et al.\egroup
  }{2015}]{Luong2015EffectiveAT}
Thang Luong, Hieu Pham, and Christopher~D. Manning.
\newblock Effective approaches to attention-based neural machine translation.
\newblock In {\em EMNLP}, 2015.

\bibitem[\protect\citeauthoryear{Marcheggiani and
  Titov}{2017}]{marcheggiani2017encoding}
Diego Marcheggiani and Ivan Titov.
\newblock Encoding sentences with graph convolutional networks for semantic
  role labeling.
\newblock In {\em EMNLP}, 2017.

\bibitem[\protect\citeauthoryear{Meng \bgroup \em et al.\egroup
  }{2018}]{Meng2017TowardsNS}
Zhao Meng, Lili Mou, and Zhi Jin.
\newblock Towards neural speaker modeling in multi-party conversation: The
  task, dataset, and models.
\newblock In {\em AAAI}, 2018.

\bibitem[\protect\citeauthoryear{Qian \bgroup \em et al.\egroup
  }{2017}]{Qian2017AssigningPT}
Qiao Qian, Minlie Huang, Haizhou Zhao, Jingfang Xu, and Xiaoyan Zhu.
\newblock Assigning personality/identity to a chatting machine for coherent
  conversation generation.
\newblock {\em arXiv preprint arXiv:1706.02861}, 2017.

\bibitem[\protect\citeauthoryear{Sabour \bgroup \em et al.\egroup
  }{2017}]{sabour2017dynamic}
Sara Sabour, Nicholas Frosst, and Geoffrey~E Hinton.
\newblock Dynamic routing between capsules.
\newblock In {\em NIPS}, pages 3859--3869, 2017.

\bibitem[\protect\citeauthoryear{Serban \bgroup \em et al.\egroup
  }{2016}]{serban2016building}
Iulian~V Serban, Alessandro Sordoni, Yoshua Bengio, Aaron Courville, and Joelle
  Pineau.
\newblock Building end-to-end dialogue systems using generative hierarchical
  neural network models.
\newblock In {\em Thirtieth AAAI Conference on Artificial Intelligence}, 2016.

\bibitem[\protect\citeauthoryear{Serban \bgroup \em et al.\egroup
  }{2017}]{Serban2017AHL}
Iulian Serban, Alessandro Sordoni, Ryan Lowe, Laurent Charlin, Joelle Pineau,
  Aaron~C. Courville, and Yoshua Bengio.
\newblock A hierarchical latent variable encoder-decoder model for generating
  dialogues.
\newblock In {\em AAAI}, 2017.

\bibitem[\protect\citeauthoryear{Shang \bgroup \em et al.\egroup
  }{2015}]{Shang2015NeuralRM}
Lifeng Shang, Zhengdong Lu, and Hang Li.
\newblock Neural responding machine for short-text conversation.
\newblock In {\em ACL}, 2015.

\bibitem[\protect\citeauthoryear{Sutskever \bgroup \em et al.\egroup
  }{2014}]{Sutskever2014SequenceTS}
Ilya Sutskever, Oriol Vinyals, and Quoc~V. Le.
\newblock Sequence to sequence learning with neural networks.
\newblock In {\em NIPS}, 2014.

\bibitem[\protect\citeauthoryear{Tao \bgroup \em et al.\egroup
  }{2019}]{tao2019multi}
Chongyang Tao, Wei Wu, Can Xu, Wenpeng Hu, Dongyan Zhao, and Rui Yan.
\newblock Multi-representation fusion network for multi-turn response selection
  in retrieval-based chatbots.
\newblock In {\em WSDM}, pages 267--275. ACM, 2019.

\bibitem[\protect\citeauthoryear{Wen \bgroup \em et al.\egroup
  }{2017}]{Wen2017ANE}
Tsung-Hsien Wen, Milica Gasic, Nikola Mrksic, Lina~Maria Rojas-Barahona, Pei
  hao Su, Stefan Ultes, David Vandyke, and Steve~J. Young.
\newblock A network-based end-to-end trainable task-oriented dialogue system.
\newblock In {\em EACL}, 2017.

\bibitem[\protect\citeauthoryear{Young \bgroup \em et al.\egroup
  }{2018}]{Young2017AugmentingED}
Tom Young, Erik Cambria, Iti Chaturvedi, Minlie Huang, Hao~David Zhou, and
  Subham Biswas.
\newblock Augmenting end-to-end dialog systems with commonsense knowledge.
\newblock In {\em AAAI}, 2018.

\bibitem[\protect\citeauthoryear{Zhang \bgroup \em et al.\egroup
  }{2018a}]{Zhang2017AddresseeAR}
Rui Zhang, Honglak Lee, Lazaros Polymenakos, and Dragomir Radev.
\newblock Addressee and response selection in multi-party conversations with
  speaker interaction rnns.
\newblock In {\em AAAI}, 2018.

\bibitem[\protect\citeauthoryear{Zhang \bgroup \em et al.\egroup
  }{2018b}]{Zhang2017NeuralPR}
Wei-Nan Zhang, Qingfu Zhu, Yifa Wang, Yanyan Zhao, and Ting Liu.
\newblock Neural personalized response generation as domain adaptation.
\newblock {\em World Wide Web}, pages 1--20, 2018.

\end{thebibliography}

\newpage
\appendix

\begin{table*}[t]
\renewcommand{\arraystretch}{1.2}
\footnotesize
\begin{center}
\begin{tabular}{|l|c|c|c|} 
\hline
Model & Embedding Average Score  & Embedding Greedy Score & Embedding Extrema Score \\
\hline
Seq2seq &  0.4733  +/- 0.00010 & 0.4728  +/- 0.00005 & 0.3067  +/- 0.00009   \\
Seq2seq (last utte) & 0.5969  +/- 0.00009  & 0.4751  +/- 0.00005  & 0.3185  +/- 0.00007  \\
HRED & 0.5151  +/- 0.00012  & 0.9053  +/- 0.00011  & 0.3252  +/- 0.00011  \\
\hline
GSN No-speaker (1-iter) & 0.7224  +/- 0.00004  & 1.0209  +/- 0.00008  & 0.5624  +/- 0.00005 \\
GSN No-speaker (2-iter) & 0.7318  +/- 0.00005  & 1.1199  +/- 0.00010  & 0.5796  +/- 0.00005  \\
GSN No-speaker (3-iter) & 0.7244  +/- 0.00004  & 0.9495  +/- 0.00006  & 0.5630  +/- 0.00004  \\
\hline
GSN W-speaker (1-iter) & 0.7133  +/- 0.00005  & 1.1193  +/- 0.00011  & 0.5860  +/- 0.00005  \\
GSN W-speaker (2-iter) & 0.7695  +/- 0.00004  & 1.0400  +/- 0.00006  & 0.6508  +/- 0.00005   \\
GSN W-speaker (3-iter) & 0.7473  +/- 0.00006  & 1.1610  +/- 0.00012  & 0.5890  +/- 0.00006 \\
\hline
\end{tabular}
\end{center}
\caption{Experimental results measured using three embedding-based metrics. Each score is followed by its standard deviation.}
\label{tab:auto_eval2}
\end{table*}

\begin{algorithm}[ht!] 
\small
\caption{Neural Graph Model}
\KwIn{1)\ a session; \ \ \ 2)\ alpha \textbf{$\alpha$}; \ \ \ \ \  3)\ iteration no. \textbf{$N$}} 
\KwOut{Generated response}
Represent the input session with state $\mathbf{S}$ using Eq.~\ref{eq:ns2}\;
Build state matrix $\mathbb{S}$ with $\mathbf{S}$ (Section \ref{ssec:CIFU} )\; 
\For{$i \ = \ 0; \ i \ < \ N; \ i \ += \ 1$}
{ 
  Compute $\Delta \mathbf{E}$ and $\Delta \mathbf{U}$ using Eq.~\ref{eq:ns12}\;
  // Backward Information Flow \\
    Update $\mathbb{S}$ using Eqs.~\ref{eq:ns10} and \ref{eq:ns11}\;
}
\For{$i \ = \ 0; \ i \ < \ N; \ i \ += \ 1$}
{ 
  Compute $\Delta \mathbf{E}$ and $\Delta \mathbf{U}$ using Eq.~\ref{eq:ns9}\;
  // Forward Information Flow \\
    Update $\mathbb{S}$ using Eqs.~\ref{eq:ns10} and \ref{eq:ns11}\;
}
Decode/generate response using Eq.~\ref{eq:ns13}\;
\Return{The generated response}\;
\label{alm:algorithm}
\end{algorithm}

\section{More Details about Experiments}
\subsection{Human Evaluation Instructions} 
{Human judges were asked to rate the generated responses and the ground truth (human responses) by given these instructions (initial training and test runs with the judges were also conducted):

\textbf{a)}	We evaluate based on “naturalness”, which includes 1) grammaticality, 2) fluency, and 3) rationality. To make the judgement process easier and to reduce the bias for each aspect, we use binary scores: good or bad. This means a response will be given three binary scores (good or bad) by each judge.

\textbf{b)}	If a response gets three good ratings by a judge, 4 is awarded to the response from the judge. If it gets two good’s and one bad, 3 is awarded. If it gets two bad’s and one good, it gets 2. If a response gets three bad’s, 1 is given. For those responses with the score of 1, the judge is further requested to separate them into two groups if one group is clearly better than the other. Each response in the better group gets the final score of 1, and each response in the other group gets the final score of 0.}
\subsection{Detailed Results in Embedding-based Metrics for Automated Evaluation}
Experimental results measured using three embedding-based metrics under different settings for different methods with sequential data and graph data are given in Table \ref{tab:auto_eval2}. 'Seq2seq (last utte)' model is trained by only using the last utterance before the final response as the input (all utterances before it are ignored). `$n$-iter' means the results are obtained after $n$ iterations. `No-speaker' is our GSN model without speaker information flow while `W-speaker' has it.

\subsection{Detailed Comparison with GCN}
Prior to the proposed GSN, Graph Convolutional Networks (GCN)~\cite{kipf2017semi} encode the graph structure directly using a neural network and train on a supervised target for all nodes with labels, thereby avoiding explicit graph-based regularization in the loss function. {The main difference between GSN and GCN is the information integration mechanism. GSN uses gated updating (calculating the update directly) while GCN uses propagation rules (estimating the output after updating). GSN uses dynamic iteration to consider long distance dependency while GCN uses multi-layers. We tested GCN by using a 3-layers GCN (corresponding to 3 iterations in GSN; 1 and 2-layers are poorer. There is a little change from 4-layers for GCN (we use only one layer). The GCN code is from https://github.com/tkipf/gcn. The experimental results for the six metrics in Table 2 are 11.86, 4.33, 2.39, 1.54, 4.36, 9.86, which are poorer than GSN. }

\section{Case Study}

\begin{table}[h]
\renewcommand{\arraystretch}{1.2}
\newcommand{\tabincell}[2]{\begin{tabular}{@{}#1@{}}#2\end{tabular}}
\scriptsize
\begin{center}
\begin{tabular}{|p{0.45\columnwidth}|p{0.45\columnwidth}|}
\hline
\tabincell{p{0.45\columnwidth}}{\textbf{Utterance 1}: my memory card for my phone won't allow me to delete files or put files on it, it says ``read only file system''. \\ \textbf{Human}: have you tried---cd filepath ls -l? \\ \textbf{No-speaker (1-iter)}: please don't flood, use url to paste, don't use enter as punctuation. \\ \textbf{No-speaker (3-iter)}: what is the output of: uname -r?\\ \textbf{W-speaker (1-iter)}: does it show up in the terminal? \\ \textbf{W-speaker (3-iter)}: you can mount it with the mount command. \\ \textbf{Seq2seq}: try that. i'm not sure how to do it. \\ \textbf{HRED}: that's what i'm looking for.} &  \tabincell{p{0.45\columnwidth}}{\textbf{SESSION}: \\ \textbf{utterance 1:} i'm running on amd athlon 64 4000. \\ \textbf{utterance 2:} then it will support 64bit if you wish to use it. \\ \textbf{utterance 3:} i understand that, but how i make .txt files not unknown?\\ \textbf{utterance 4:} should i use it? whats the difference? \\ \textbf{RESPONSES}: \\ \textbf{No-speaker (3-iter)}: i'm not sure what you're looking for.\\ \textbf{W-speaker (1-iter)}: in a terminal, type ``sudo apt-get update'' \\ \textbf{W-speaker (3-iter)}: you need to install it from source. \\ \textbf{Seq2seq}: i'm not sure what you ' re trying to do. \\ \textbf{HRED}: you ' ll need to find out what you ' re doing .} \\
\hline
\end{tabular}
\end{center}
\caption{Sample responses generated by human, GSN, and baselines. All the responses in column 1 are for utterance 1. Column 2 is a session with the graph structure in Figure 5.}
\label{tab:casestudy}
\end{table}

Table \ref{tab:casestudy} presents some sample responses generated by our model and the baselines. From the first example, we see that the responses generated by our model are more consistent with the given utterance.
The utterance in the first example (on the left in Table \ref{tab:casestudy}) is asking for help, not offering an answer. However, HRED mistakenly takes it as a useful answer and gives a response ``that's what i'm looking for'', which may not be appropriate. The seq2seq model provides a meaningless utterance "try that. i'm not sure how to do it", which addresses nothing. In contrast to these methods, our model tries to give some useful suggestions, e.g., ``you can mount it with the mount command.'' and ``what is the output of: uname -r?'', etc, which we believe are more effective. 

In the second column (Table \ref{tab:casestudy}), based on utterances $1$, $2$ and $3$, the speaker tried to run a 64-bit application but the application failed with an error message ``.txt files unknown''. In this case, our models generate meaningful responses from the training corpus: giving a suggestion to update the system (W-speaker 1-iter); advising the speaker to try another way to install the software (W-speaker 3-iter). Thus, our models (with the speaker information flow) generate responses relevant to the entire session and give substantive advice for the raised questions.
These responses are more appropriate and preferred from the human perspective.  

\end{document}